\documentclass[conference]{IEEEtran}
\IEEEoverridecommandlockouts
\usepackage{amsmath,amssymb,amsfonts}
\usepackage{algorithmic}
\usepackage{graphicx}
\usepackage{textcomp}
\usepackage{xcolor}
\usepackage{threeparttable}
\usepackage[style=ieee]{biblatex}
\bibliography{source.bib}
\def\BibTeX{{\rm B\kern-.05em{\sc i\kern-.025em b}\kern-.08em
    T\kern-.1667em\lower.7ex\hbox{E}\kern-.125emX}}
\begin{document}

\title{Analyzing the Effectiveness of Image Augmentations for Face Recognition from Limited Data}

\author{\IEEEauthorblockN{Aleksei Zhuchkov}
\IEEEauthorblockA{\textit{Innopolis University} \\
Innopolis, Russia \\
a.zhuchkov@innopolis.ru}
}

\maketitle

\begin{abstract}
This work presents an analysis of the efficiency of image augmentations for the face recognition problem from limited data. We considered basic manipulations, generative methods, and their combinations for augmentations. Our results show that augmentations, in general, can considerably improve the quality of face recognition systems and the combination of generative and basic approaches performs better than the other tested techniques.
\end{abstract}


\section{Introduction}
\label{chap:intro}
In the past two decades, artificial neural networks(ANNs) have achieved impressive performance in numerous applications, including object recognition~\cite{khan2020post}, anomaly detection~\cite{rivera2020anomaly, yakovlev2020abstraction}, accident detection~\cite{batanina2019domain, bortnikov2019accident}, action recognition~\cite{gavrilin2019across, sozykin2018multi, khan2010accelerometer}, scene classification~\cite{protasov2018using}, hyperspectral image classification~\cite{ahmad2020fast, ahmad2019multi}, medical image analysis~\cite{gusarev2017deep, dobrenkii2017large}, machine translation~\cite{khusainova2019sart, valeev2019application}, and face recognition (FR)~\cite{he2020pa}. However, the quality and amount of input data significantly affect the performance of these methods. Collecting and labeling a large amount of face image samples is expensive, labor-intensive, and error-prone. Small companies in the IT industry often lack access to huge corpora of data. For example, the current state-of-art FR technique FaceNet with adaptive threshold~\cite{chou2019data} has high accuracy results. However, it was trained on MS-Celeb-1M dataset \cite{guo2016ms} that is quite large.

Data augmentation is a technique that expands the amount and diversity of training data by applying various transformations to images
\cite{khan2020post}, making them a potential solution to limited-data face recognition problem.  However, there is a gap in the scientific literature considering research on augmentations’ theoretical and methodological aspects for FR. Although some works tried to solve the problem of making decisions related to the number, type, and intensity of augmentations \cite{sato2015apac}, only a small number of papers performed a rigorous comparison of the different augmentation techniques. Moreover, there is an overall lack of work on the expansion of the size of small datasets using augmentations. Accordingly, in this work, we aim to fill this gap by comparing various augmentation techniques for FR from limited data. 
    
The rest of the paper is organized as follows. Section \ref{chap:lr} provides an overview of the related work. Section \ref{chap:met} describes the research methodology we used. Section \ref{chap:res} summarizes the findings of the study and explains how they could be interpreted. Finally,
Section \ref{chap:con} makes the conclusion and proposes interesting
directions for future research.

\section{Related Work}
\label{chap:lr}
Image augmentations can be divided into two broad categories: traditional augmentation techniques and generative. Traditional augmentation techniques (also referred to as basic transformations)
include a variety of techniques e.g. geometric transformations, random crop, and color space transformations. Most common geometric transformations include flipping, cropping, rotation, translation, and noise injection \cite{noh2017regularizing, kwasigroch2017deep}.
These methods are easy to understand, implement, and reproduce.  However, despite the popularity of geometric data augmentations, they can be applied in a limited scope and they are still the subject of research. Recently, \cite{zhong2020random}
showed that random erasing can make ANNs more robust to different kinds of defects on images. Random erasing is easy to implement, parameter learning free, and can be combined with different kinds of data augmentations.
Color space transformations are related to the manipulation of color values of an image \cite{galdran2017data}.  Their limitations include increased training time and higher memory utilization.

Generative approaches include such methods as Neural Style Transfer (NST) and Generative adversarial networks (GAN).  NST \cite{zheng2019stada} synthesizes a new image using style from a style-image and content from content-image. It facilitates the transfer of textures, color temperatures, and lighting conditions, but it can lead to bias in the dataset. In addition, original NST algorithms \cite{gatys2016image} are slow.
GANs \cite{goodfellow2014generative} create samples similar to images from the source dataset. GAN data augmentation approach can boost the performance of the model even if generated samples do not look hyper-realistic \cite{chaitanya2019semi}, making them suitable for medical problems \cite{xue2019synthetic}. 
The main problem of GANs is their unstable training and consumption of a considerable amount of training data, making them impractical for limited-data problems.

Nowadays, deep learning approaches are mainstream methods for face recognition tasks. The most famous neural nets used for face recognition tasks which achieved promising results on the LFW dataset are DeepFace \parencite{taigman2014deepface}, DeepID \parencite{sun2014deep} and FaceNet \parencite{schroff2015facenet}. FaceNet's authors suggested a new loss known as the Triplet Loss. It reduces the distance between an anchor image and a positive example of the same identity and increases the distance between the anchor and a negative example with a different identity. The current state-of-the-art method is also based on the CNN \parencite{chou2019data} and has two operations in the system: registration and recognition. In the operation of registration, an embedding is extracted from an input face image by using a FaceNet model. The threshold is assigned to the registered face during each registration (stored in the system), and the thresholds of the other stored faces are modified accordingly in the system. Recognition part is described in details in Section \ref{sec:eval_met}.

\section{Methodology}
\label{chap:met}

As stated earlier, this work explores the performance of different data augmentation techniques (basic, generative, and their combination) for limited-data face recognition. Basic manipulations are simple and easy to perform, but they cannot show realistic face variations. On the other hand, the generative approach can produce great realistic face variations but requires extra resources. To answer our research questions we decided to split our research into five main steps to investigate this problem and find the best approach for augmentation.

\begin{enumerate}
 
    \item Determine an approach for the realistic generation of faces with different attributes such as different hair colors or glasses.
    \item Determine the strategy for image augmentations using basic approaches e.g. geometric, random occlusion, or color jitter.
    \item Generate datasets using only basic manipulations, only generative approach, and the combination of these two methods.
    \item Choose a state-of-the-art face recognition model architecture.
    \item Train the chosen recognition model using generated datasets and measure its performance.
 
\end{enumerate}

\subsection{Dataset}
\label{sec:dataset}
We chose the Labeled Faces in the Wild (LFW) dataset \parencite{LFWTech} for training and testing in our research. It is a collection of face images created to study the issue of unconstrained face recognition.  After the LFW dataset cleaning, it consists of 13156 images of faces found on the internet. Overall, there are 5718 different people in LFW. Among them, most are presented by less than 10 distinct images. Thus, this dataset can be considered small and fits our research needs. We decided to split the dataset into two different ways to produce two different training datasets. In the first split, all images of one person are either in the train or the test dataset. We called it the Unique split. It is used to test the generalization abilities of models. In the second split, if there is more than one image of a person, then one of the pictures is assigned to test and the rest belongs to the train split. We called it the Both split. If there is only one picture of a person, this image is placed into the training dataset. Each split has a train/test proportion close to 0.9/0.1.

\subsection{Realistic face attributes generation}
\label{sec:realistic}
As a realistic approach, we used a PA-GAN \parencite{he2020pa} to generate attributes on faces. It is a generative approach and shows good performance in generating different face attributes such as glasses, hair of a different color, or smiles. PA-GAN model requires alignment of a dataset based on predefined landmarks. We used Dlib library \footnote{http://dlib.net/} to detect landmarks on LFW dataset. Then we used an affine transformation to transform face image from original to predefined landmarks. 

For facial attribute editing this approach uses a progressive attention GAN that edits the attributes in an attention-based manner as they progress from high to low feature stages. The attribute editing is carried out in a coarse-to-fine way, which is stable and accurate, due to the progressive focus process. The authors of PA-GAN conducted experiments on the CelebA dataset and their method achieved state-of-the-art results.

\subsection{Image generation using basic transformations}
\label{sec:basic_trans}
We used several basic transformations to simulate changes that can happen with a face (accessories, beard, hair color, skin tone, and more). For example, random black occlusion can simulate wearing a beard or glasses. Grid distortion can change the width or height of the face which can happen to a person if he or she loses or gains some weight. Moreover, changing image parameters such as brightness, saturation or contrast can simulate skin and hair tone changing.

\subsection{Generating datasets}
\label{sec:dataset_gen}
We generated 24 augmented images for each original image in the training datasets in each of the six produced datasets (generative, combined, and basic approaches for each of 2 training splits).

\subsubsection{Generative}
We used the official implementation \footnote{\url{https://github.com/LynnHo/PA-GAN-Tensorflow}} for attributes generation on the LFW dataset. It already has a pre-trained model that can generate the following features:

\begin{itemize}
 
  \item age: old, young
  \item bangs: yes, no
  \item eyebrows: usual, bushy
  \item eyeglasses: yes, no
  \item gender: male, female
  \item facial hair: beard, mustache, no hair
  \item hair: bald, blond, black, brown
  \item mouth: open, closed
  \item skin: pale, usual
  
\end{itemize}

\begin{figure}
\leavevmode
\hbox to\linewidth{%
    \hfil%
    \vbox{%
        \hbox{\includegraphics[width=.18\linewidth]{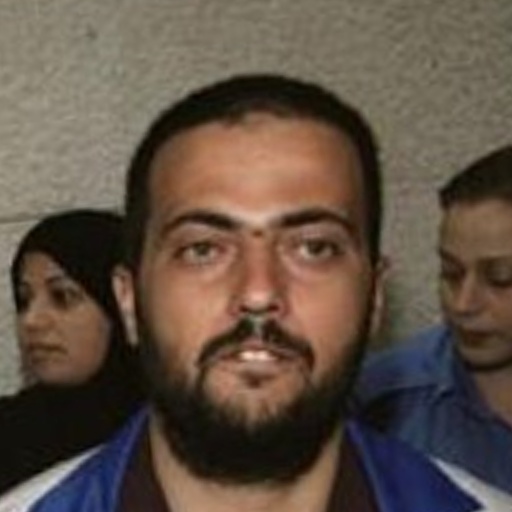}}
    }%
    \hfil%
    \vbox{%
        \hbox{\includegraphics[width=.18\linewidth]{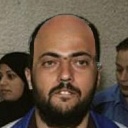}}
    }%
    \hfil%
    \vbox{%
        \hbox{\includegraphics[width=.18\linewidth]{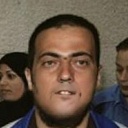}}
    }%
    \hfil%
    \vbox{%
        \hbox{\includegraphics[width=.18\linewidth]{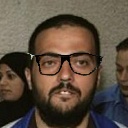}}
    }%
    \hfil%
    \vbox{%
        \hbox{\includegraphics[width=.18\linewidth]{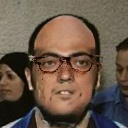}}
    }%
    \hfil%
}
\caption{Example of changing attributes on aligned image from LFW dataset using PA-GAN.}
\label{fig:pa_gan_lfw}
\end{figure}

We generated all combinations of hair (4 options), eyeglasses (2 options) and hair face (3 options) attributes that resulted in 24 images. All other attributes were added to combinations randomly. Fig. \ref{fig:pa_gan_lfw} shows the examples of images from aligned LFW dataset; the images were pre-processed by PA-GAN.  

\subsubsection{Basic manipulations}
\begin{figure}
\leavevmode
\hbox to\linewidth{%
    \hfil%
    \vbox{%
        \hbox{\includegraphics[width=.18\linewidth]{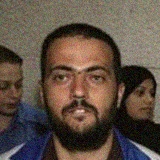}}
    }%
    \hfil%
    \vbox{%
        \hbox{\includegraphics[width=.18\linewidth]{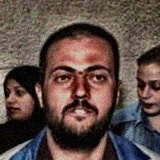}}
    }%
    \hfil%
    \vbox{%
        \hbox{\includegraphics[width=.18\linewidth]{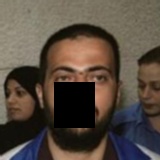}}
    }%
    \hfil%
    \vbox{%
        \hbox{\includegraphics[width=.18\linewidth]{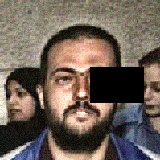}}
    }%
    \hfil%
    \vbox{%
        \hbox{\includegraphics[width=.18\linewidth]{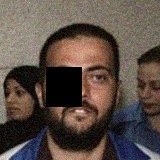}}
    }%
    \hfil%
}
\caption{Example of basic manipulations application on aligned images from LFW dataset}
\label{fig:basic_lfw}
\end{figure}

We used Albumentations library \footnote{https://albumentations.ai/} for applying basic manipulations on train images. Fig. \ref{fig:basic_lfw} shows the example of application of basic manipulations: random black occlusion, changing brightness, contrast, and saturation parameters, adaptive histogram equalization, blur and downscale of the image, horizontal flip, random distortion, Gauss noise, and grid distortion.

\subsubsection{Combining by consecutive applying}
\begin{figure}
\leavevmode
\hbox to\linewidth{%
    \hfil%
    \vbox{%
        \hbox{\includegraphics[width=.18\linewidth]{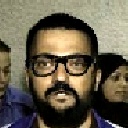}}
    }%
    \hfil%
    \vbox{%
        \hbox{\includegraphics[width=.18\linewidth]{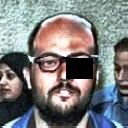}}
    }%
    \hfil%
    \vbox{%
        \hbox{\includegraphics[width=.18\linewidth]{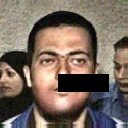}}
    }%
    \hfil%
    \vbox{%
        \hbox{\includegraphics[width=.18\linewidth]{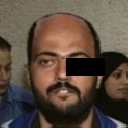}}
    }%
    \hfil%
    \vbox{%
        \hbox{\includegraphics[width=.18\linewidth]{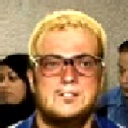}}
    }%
    \hfil%
}
\caption{Examples of images after application of basic manipulations. Source images are produced by the generative approach from aligned images from the LFW dataset.}
\label{fig:combined_lfw}
\end{figure}

The central idea of this method is to combine strategies. We applied basic manipulations on images produces by the generative approach. The number of augmented images did not change: it was still 24 images per face. Fig. \ref{fig:combined_lfw} shows the examples of augmented images using a combined technique.

\subsection{Choice of face recognition model}
\label{sec:face_rec_model}
As the face recognition approach, we chose the FaceNet model with adaptive threshold. It shows near to state-of-the-art performance on our reference LFW dataset \parencite{chou2019data}. We used the official implementations of FaceNet \footnote{\url{ https://github.com/davidsandberg/facenet}} and Adaptive threshold \footnote{\url{https://github.com/ivclab/Online-Face-Recognition-and-Authentication}} to construct our experiments. However, we did not use a pre-trained FaceNet model because our goal is to simulate a case when only a small amount of data is available.

\subsection{Training and testing of the models}
\label{sec:train_test}
After generating datasets we trained the FaceNet model on different datasets resulting in eight models. Two of them are trained on the Unique and Both train splits. We trained six models on generated datasets that were produced using the methods described above. Then, using the Adaptive threshold technique we measured the performance of trained models in the face recognition task on test splits of the LFW dataset.

\section{Evaluation and Experiment}
\label{chap:res}

\subsection{Evaluation metrics}
\label{sec:eval_met}

\begin{figure}[!htb]
    \includegraphics[width=\linewidth]{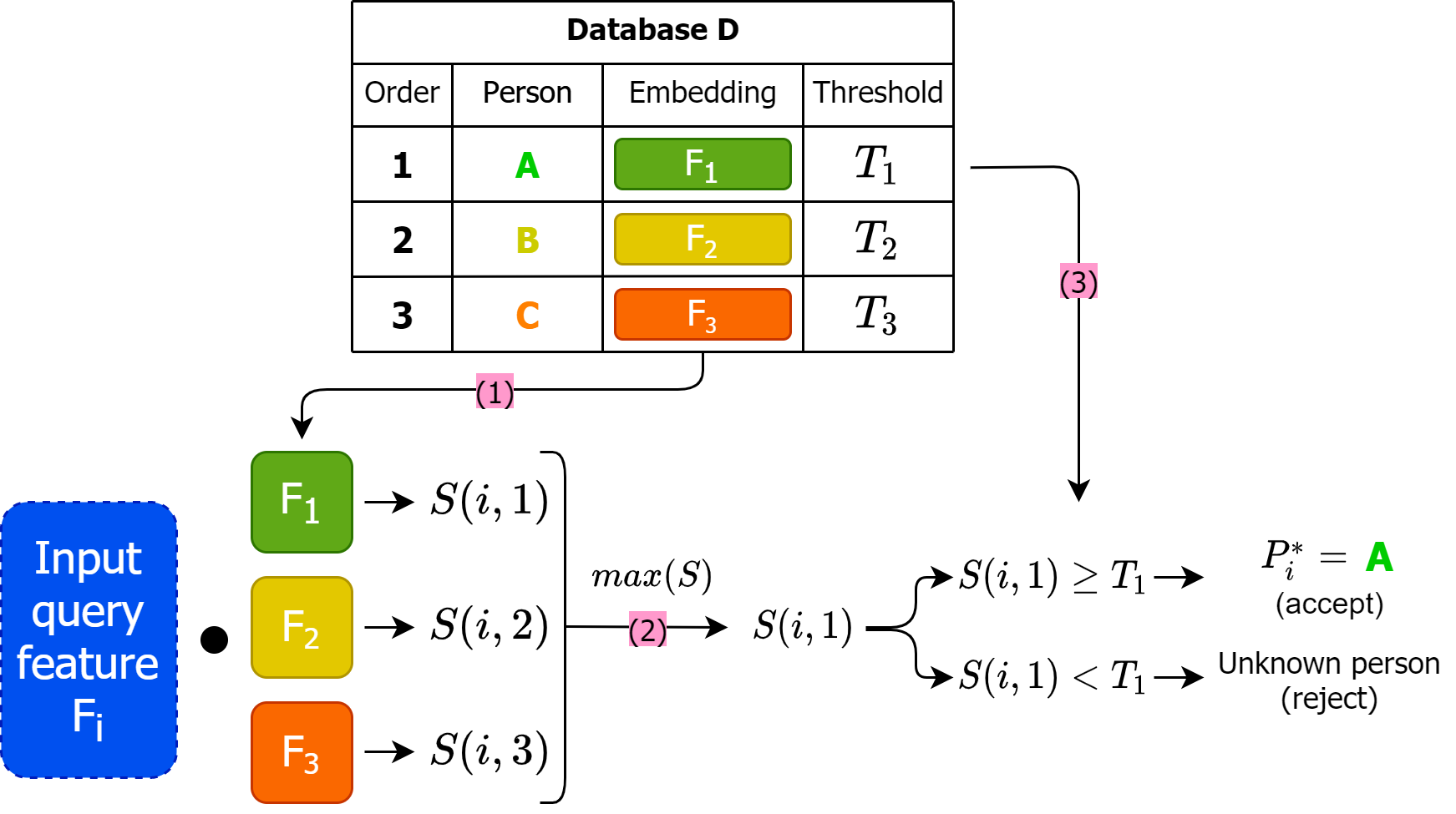}
    \caption{Recognition pipeline. $F_i$ is an embedding of $I_i$ image from the test dataset. (1) Compute similarity scores $S(i, j)$ of each of embeddings $F_j$ from database with input $F_i$. (2) Get the maximum similarity score $S(i, m)$, where m is the order of embedding with the highest similarity score in the database. As an example, $S(i, 1)$ has been taken as maximum on the diagram. (3) Compare maximum score $S(i, m)$ with the threshold $T_m$ stored in the database for embedding $F_m$. If the score is greater than the threshold, then the input image is accepted and assigned embedding $F_m$'s label. $P_i^*$ is assigned label for $F_i$. Otherwise, the input image is rejected as an unknown person.}
    \label{fig:rec_scheme}
\end{figure}

Before going into details about the models' results, it is necessary to describe the metrics to compare models. The most intuitive efficiency metric is accuracy, which is in general the number of correct observations to all observations. We chose a slightly modified accuracy metric that was introduced in the work of \citeauthor{chou2019data} \cite{chou2019data} and will be described below. An experiment starts with a test dataset of size N that consists of images $\boldsymbol{I}= \{I_1, I_2, ..., I_N\}$ and labels $\boldsymbol{P}= \{P_1, P_2, ..., P_N\}$. The system has an empty database D, which is used to store information about people known to the system, feature vectors obtained from their images, and thresholds $T$. For each image $I_i$, feature vector $F_i$ is calculated using FaceNet. Then, the system calculates similarity scores  $S(i, j)$ for each $F_i$ to each vector $F_j \in D$ where $0 <  j < |D|$. Afterwards, the system determines the maximum score $S(i, m)$ out of calculated similarity scores, where $m$ is the order of embedding $F_m$ with the highest score in the database. Next, the system compares the score $S(i, m)$ to threshold $T_m$ associated with $F_m$ from the database and predicts label $P_i^*$ for $F_i$. After that, predicted label $P_h^*$ is compared to the true label $P_i$. Tab. \ref{table:acc_for} shows the protocol which is used for evaluation of predictions. Fig. \ref{fig:rec_scheme} describes the whole recognition process for image $I_i$. Finally, the image vector $F_i$ and its label $P_i$ are saved in the database after the comparison stage. The whole process repeats for the next image $I_{i+1}$. The final accuracy is defined as the average correctness of predictions for all N images as shown in Eq. \ref{eq:acc}, where TA is true accept and TR is true reject.

\begin{table}[h!]
\centering
    \begin{threeparttable}
        \caption{Evaluation protocol of person predictions}
        \begin{tabular}{ |c|c|c| } 
            \hline
              & $S(i, m) \geq T_m$ & $S(i, m) < T_m$\\
            \hline
            $P_i^* = P_i$  & True accept (TA) & False reject (FR)\\
            \hline
            $P_i^* \neq P_i$ and $P_i^* \in D$ & Identification error (IE)& False reject (FR)\\
            \hline
            $P_i^* \neq  P_i$ and $P_i^* \notin D$ & False accept (FA) & True reject (TR) \\
            \hline
        \end{tabular}
        \begin{tablenotes}
          \small
          \item Note: $F_i$ is an embedding of $I_i$ image from the test dataset. $P_i$ is the true label of $F_i$. $P_i^*$ is assigned label for $F_i$ by recognition pipeline. $S(i, m)$ represents maximum similarity that is between $F_i$ and $F_m$. $T_m$ is the threshold for $F_m$ stored in the database.
        \end{tablenotes}
        \label{table:acc_for}
    \end{threeparttable}
\end{table}

\begin{equation} \label{eq:acc}
Accuracy = \frac{\sum_{i=1}^N |TA(i)| + |TR(i)|} {N}
\end{equation}

As we have already mentioned, experiment starts with a test dataset of size N that consists of images $I = \{I_1, I_2, ..., I_N\}$ and labels $P= \{P_1, P_2, ..., P_N\}$. Fig. \ref{fig:rec_scheme} shows that based on a comparison of similarity score and threshold, the input image can be accepted or rejected. Thus, we have several accepted (ACP) and rejected (REJ) images when all images are passed through the system. A few metrics can be calculated based on ACP/REJ numbers and the evaluation protocol that are shown in Table \ref{table:acc_for}. These metrics can give better insights into the actual performance of face recognition models and show the strengths and weaknesses of the models.

\begin{itemize}
 
    \item True Accept Rate (TAR) is the ratio of correctly accepted images to the total number of accepted pictures. 
        \begin{equation} \label{eq:tar}
        TAR = \frac{\sum_{i=1}^N |TA(i)|} {ACP}
        \end{equation}
    
    \item True Reject Rate (TRR) is the ratio of correctly rejected images to the total number of rejected pictures.
        \begin{equation} \label{eq:trr}
        TRR = \frac{\sum_{i=1}^N |TR(i)|} {REJ}
        \end{equation}
        
    \item False Accept Rate (FAR) is the ratio of wrongly accepted images to the total number of accepted pictures.
        \begin{equation} \label{eq:far}
        FAR = \frac{\sum_{i=1}^N |FA(i)|} {ACP}
        \end{equation}
        
    \item False Reject Rate (FRR) is the ratio of wrongly rejected images to the total number of rejected pictures.
        \begin{equation} \label{eq:frr}
        FRR = \frac{\sum_{i=1}^N |FR(i)|} {REJ}
        \end{equation}
        
    \item Wrong Identification Rate (WAR) is the ratio of wrongly predicted labels for accepted images to the total number of accepted pictures.
        \begin{equation} \label{eq:war}
        WAR = \frac{\sum_{i=1}^N |IE(i)|} {ACP}
        \end{equation}
  
\end{itemize}

\subsection{Results}
\label{sec:model_res}

As mentioned in Section \ref{chap:met}, we had two test datasets that were produced by different splits of the source LFW dataset. Additionally, we had eight models of the same architecture but trained on different datasets. We conducted 10 runs of each model on a test set that corresponds to its split and took their average accuracy. We restricted calculations of adaptive threshold only to 100 entries in the database to speed up the testing phase. 

Table \ref{table:met_unique} contains the metrics obtained from the test set with Unique split. The findings on Unique split show the increase in accuracy when a combined augmentation technique is used compared to baseline results. This method shows the lowest false acceptance rate among all tested methods. The basic approach demonstrated accuracy results close to the combined approach. It shows the highest true acceptance and true rejection rates. Moreover, the basic approach has the lowest false rejection and wrong identification rates. The generative augmentation method shows results considerably worse than the basic and combined approaches and they are close to the baseline. 

It is worth taking a closer look at the classification results. Examples of misclassified images of identities on test dataset with Unique split are shown in Table \ref{table:unique_exs}. It can be noticed that in general, all models misidentified people who have similar facial features or photos that have a similar color scheme. Moreover, models show false rejection results on the same identities when photos from the database and input photo have different accessories, lightning conditions, or contrast background. However, models trained on datasets augmented by generative, combined, and basic methods are less prone to this problem.

Both split is constructed so that if there is more than one image of a person in the dataset, then one of the pictures is assigned to test and the rest belongs to the train split. Thus, the test dataset with Both split contains only one image per identity and contains identities that are in the training dataset. Based on the identity recognition scheme shown in Fig. \ref{fig:rec_scheme}, we can say that model needs to reject all input pictures to get accuracy equal to 1.0. Table \ref{table:met_both} shows metrics derived from testing models on the test dataset with Both split. Based on received information, models trained on baseline dataset and datasets generated using generative and combined approaches show similar accuracy results. However, the basic manipulations method has degraded results compared to baseline.

\begin{table}[h!]
    \centering
    \begin{threeparttable}
        \caption{The average metrics of the models which were run 10 times on test dataset with Unique split}
        \begin{tabular}{ |c|c|c|c|c|c|c| } 
            \hline
                Method & Baseline & Generative & Basic & Combined & $\mu_{unique}$\\
            \hline
                ACC  & 0.3574 & 0.3668 & 0.4628 & \textbf{0.4637} & 0.4127\\
            \hline
                TAR  & 0.1137 & 0.0687 & \textbf{0.3562} & 0.2876 & 0.2065\\
            \hline
                TRR  & 0.5550 & 0.5474 & \textbf{0.7058} & 0.6071 & 0.6038\\
            \hline
                FAR  & 0.4721 & 0.4698 & 0.4356 & \textbf{0.4087} & 0.4465\\
            \hline
                FRR  & 0.4449 & 0.4525 & \textbf{0.2941} & 0.3928 & 0.3960\\
            \hline
                WAR  & 0.4140 & 0.4614 & \textbf{0.2081} & 0.3035 & 0.3467\\
            \hline
        \end{tabular}
        \begin{tablenotes}
              \small
              \item Note: $\mu_{unique}$ is a mean of the metric for all methods.
        \end{tablenotes}
        \label{table:met_unique}
    \end{threeparttable}
\end{table}

\begin{table} 
    \centering
    \caption{Examples of misclassification of identities on test dataset with Unique split} 
    \begin{tabular}{|c|c|c|c|c|}
        \hline
            Method & \multicolumn{2}{|c|}{Wrong identification} &
            \multicolumn{2}{|c|}{False reject}\\
        \cline{2-5}
            & Database & Predicted & Database & Rejected\\
        \hline
            Baseline &
            \includegraphics[scale = 0.07]{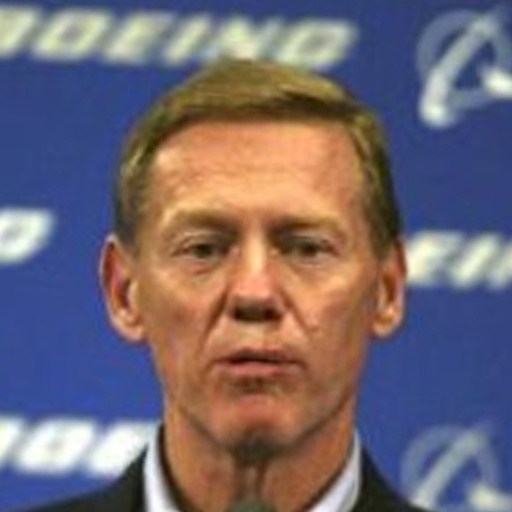} & \includegraphics[scale = 0.07]{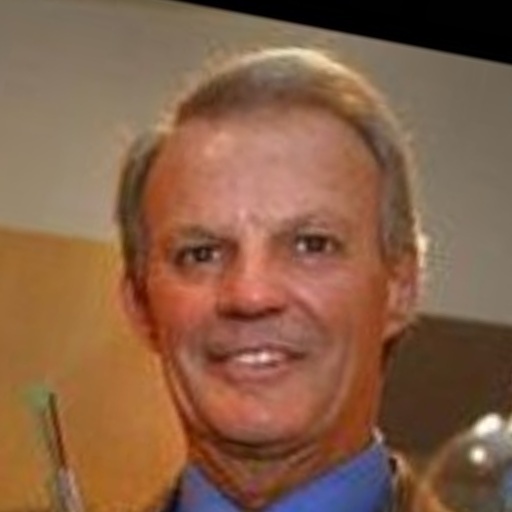} & \includegraphics[scale = 0.07]{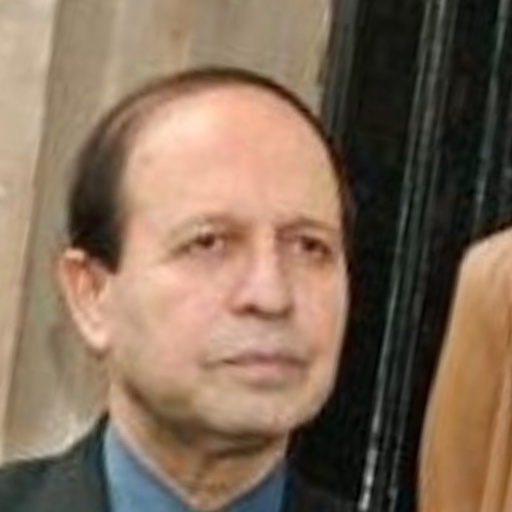} & \includegraphics[scale = 0.07]{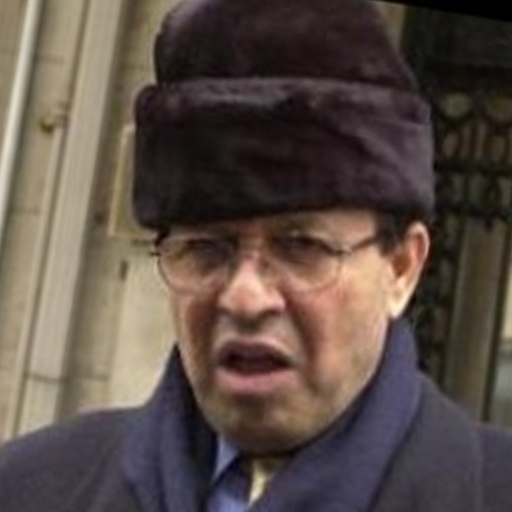}\\ 
        \hline
            Generative &
            \includegraphics[scale = 0.07]{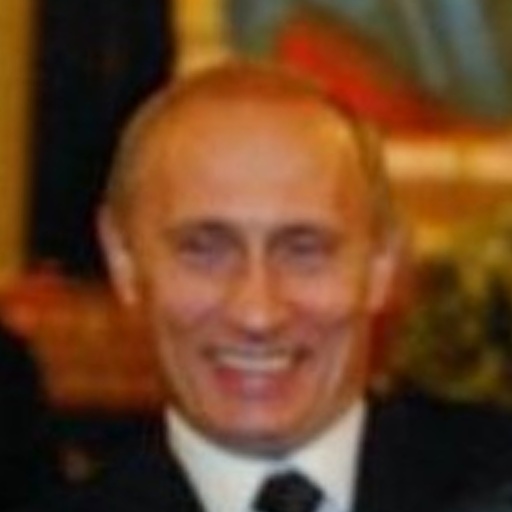} & \includegraphics[scale = 0.07]{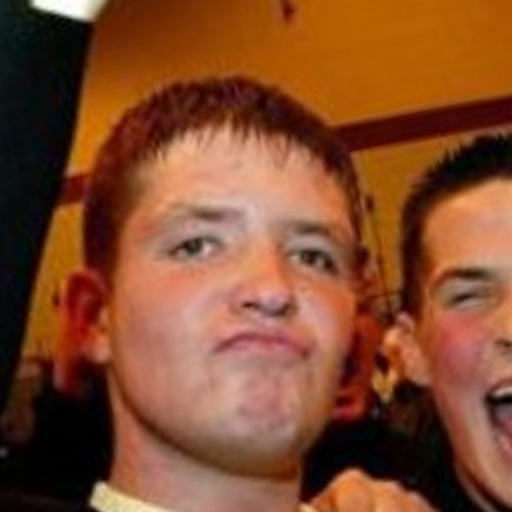} & \includegraphics[scale = 0.07]{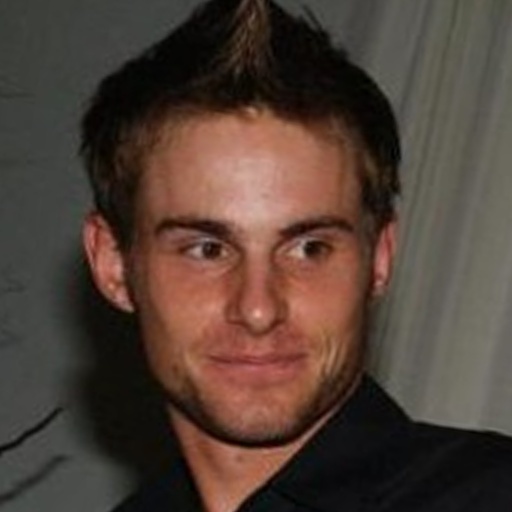} & \includegraphics[scale = 0.07]{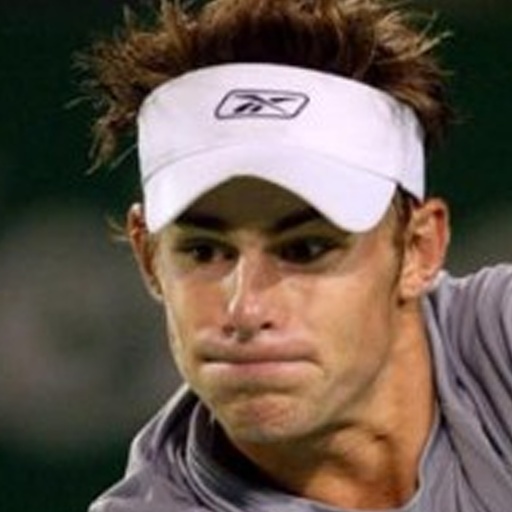}\\ 
        \hline
            Basic &
            \includegraphics[scale = 0.07]{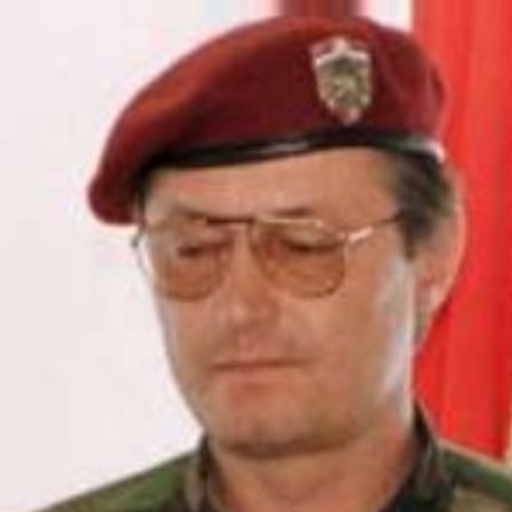} & \includegraphics[scale = 0.07]{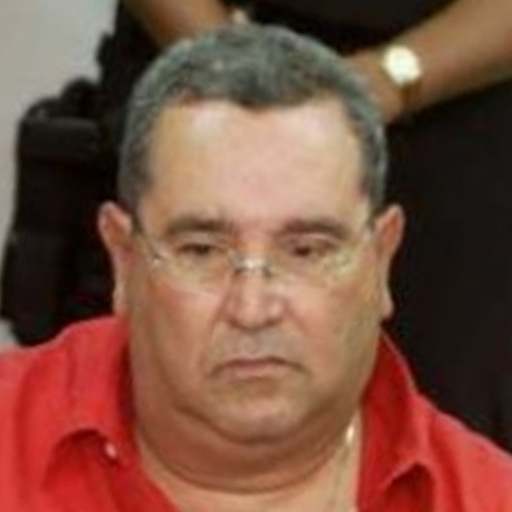} & \includegraphics[scale = 0.07]{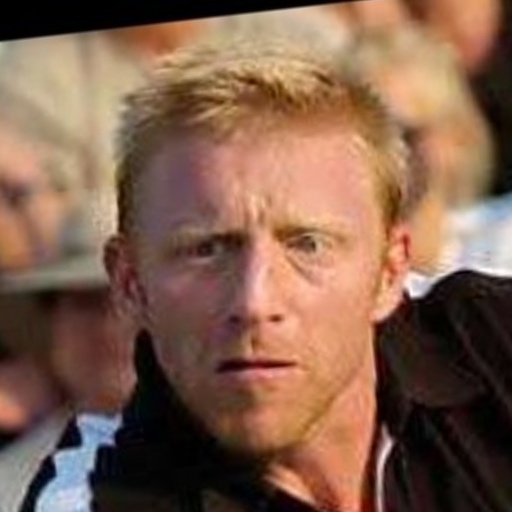} & \includegraphics[scale = 0.07]{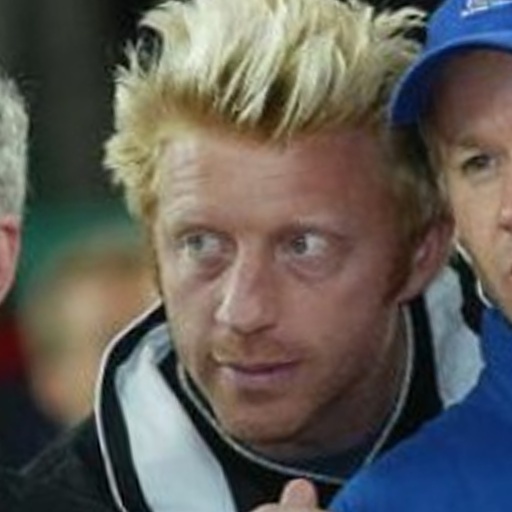}\\ 
        \hline
            Combined &
            \includegraphics[scale = 0.07]{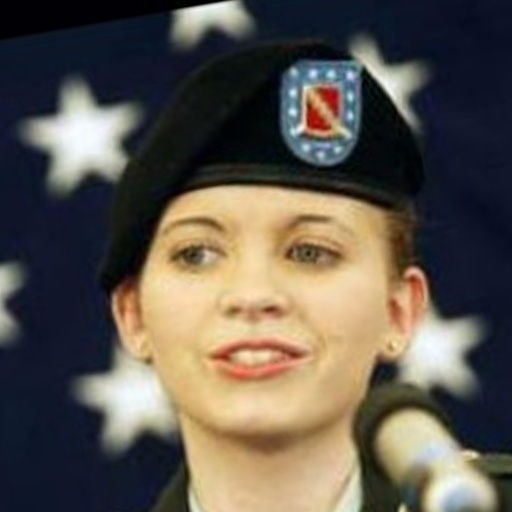} & \includegraphics[scale = 0.07]{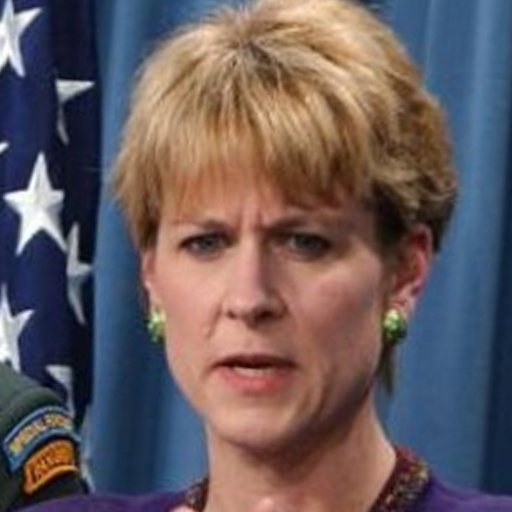} & \includegraphics[scale = 0.07]{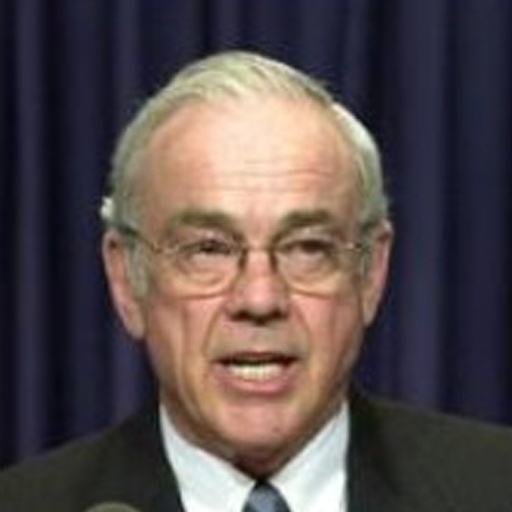} & \includegraphics[scale = 0.07]{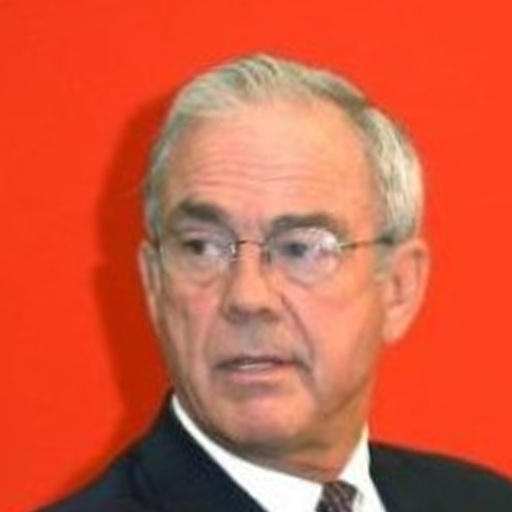}\\ 
        \hline
    \end{tabular} 
\label{table:unique_exs} 
\end{table}

\begin{table}[h!]
    \centering
    \caption{The average metrics of the models passed 10 times on test dataset with Both split}
    \begin{tabular}{ |c|c|c|c|c|c|c| } 
        \hline
            Method & Baseline & Generative & Basic & Combined & $\mu_{both}$\\
        \hline
            ACC  & 0.6062 & 0.6095 & 0.3632 & \textbf{0.6169} & 0.5489\\
        \hline
            TAR  & 0.0000 & 0.0000 & 0.0000 & 0.0000 & 0.0000\\
        \hline
            TRR  & 1.0000 & 1.0000 & 1.0000 & 1.0000 & 1.0000\\
        \hline
            FAR  & 1.0000 & 1.0000 & 1.0000 & 1.0000 & 1.0000\\
        \hline
            FFR  & 0.0000 & 0.0000 & 0.0000 & 0.0000 & 0.0000\\
        \hline
            WAR  & 0.0000 & 0.0000 & 0.0000 & 0.0000 & 0.0000\\
        \hline
    \end{tabular}
    \label{table:met_both}
\end{table}

\subsection{Discussion}
\label{sec:discussion}

This work introduces novel research on measuring the performance of augmentations for face recognition problems. Experiments were conducted to assess the performance of models trained on different datasets. In general, augmentations increased the performance of face recognition models. Results showed a significant increase in accuracy when the combined augmentation technique was used in the Unique split and a slight increase in Both split cases. Nevertheless, the basic approach demonstrated results close to the combined approach for the Unique split case. It means that the generative approach did not make a big contribution to the increase of accuracy of the combined approach. However, the basic approach caused a great accuracy degradation in Both split case. Meanwhile, the generative approach showed a little improvement compared to the baseline. Experiments show that the basic approach contributes to the increase of the true acceptance rate and decrease of misidentifications of identities. At the same time, the generative approach works well with the increase of the true rejection rate. A combination of approaches probably contributes to their strengths to deliver better results. Results suggest that using a combined approach can help to overcome the problem of a small training dataset. However, the basic approach can be preferable since it showed quite good results in an unconstrained unique split experiment and since it requires fewer resources than the generative one.

\section{Conclusion}
\label{chap:con}
The main problem for face recognition systems is diversity in the intra‐subject face's images. This diversity can be caused by absence of structuring elements or the presence of components such as a beard and/or a mustache, a cap, sunglasses, etc., or occlusions of the face by background or foreground objects. The goal of the current research was to measure the effectiveness of two big classes of augmentations - basic manipulations and generative approaches on small datasets in face recognition problems and compare their performance. Thus, we generated several datasets using the two approaches mentioned above and their combination. Then we trained several face recognition models and achieved accuracy results on the test datasets. The experiments showed that augmentations can significantly boost the performance of face recognition systems. Moreover, although the combinations of augmentation approaches demonstrated a good accuracy increase, the basic approach is not far behind in performance but demands less time and hardware resources. 

Several enhancements can be done for further development of the researched topic. Firstly, another dataset with more various examples can be taken. Diverse datasets for training lead to a better generalized model. Another way of research is to try more complex augmentations approaches, e.g that change face rotation angle, add more accessories, or change the background.

\printbibliography
\end{document}